\documentclass[cameraready]{Interspeech}

\usepackage[]{todonotes}
\makeatletter
\newcommand*\iftodonotes{\if@todonotes@disabled\expandafter\@secondoftwo\else\expandafter\@firstoftwo\fi}  

\everypar{\looseness=-1}

\title{What Counts as an Error? Dual-Reference Benchmarking for Atypical ASR}

\author[affiliation={1}, orcid=0009-0008-3927-526X]{Hawau Olamide}{Toyin}
\author[affiliation={2}]{Srinivasan}{Umesh}
\author[affiliation={1}, orcid=0000-0003-1706-1777]{Hanan}{Aldarmaki}

\address{
  $^1$MBZUAI, UAE; $^2$SPRING Lab,
IIT Madras, India}

\email{\{hawau.toyin,hanan.aldarmaki\}@mbzuai.ac.ae}


\usepackage{comment}
\usepackage{booktabs}
\usepackage{multirow}
\usepackage[table,xcdraw]{xcolor}
\usepackage{tabularx}
\usepackage{array}

\begin{document}

\maketitle

\begin{abstract}

    ASR systems have been often reported to underperform on atypical speech. An often conflated compounding factor is the existence of two valid transcription references: verbatim (actual produced speech, including repetitions/prolongations) and intended (the canonical form of the text with disfluencies removed) in atypical speech recognition depending on context and use-case. Most ASR evaluations conflate this duality into a single ground truth and reward systems that delete disfluencies, ignoring verbatim faithfulness. We benchmark 11 ASR models from encoder-decoder, CTC and transducer families using both verbatim and intended references on atypical stuttered speech as a case study. Our quantitative assessment underlines the disparity in model performance and rankings using the two transcript styles. Through this analysis, we highlight the importance of selecting a suitable transcription reference for valid model selection depending on the use-case, particularly for atypical ASR.

   
\end{abstract}

\section{Introduction}

Atypical speech of persons who stutter (PWS) includes disfluencies (e.g. repetitions, prolongations, interjections) with variable frequencies depending on the stuttering severity level. Prior works document that current ASR and voice assistant systems often fail for such atypical speech, with failure rates increasing as severity increases \cite{user-perceptions-technical,Mitra2021AnalysisAT,Mujtaba2025FineTuningAF}. For instance, they often produce transcripts that distorts what was actually said \cite{Sridhar2025JjjjustSB}, fail at voice assistant commands~\cite{user-perceptions-technical}, and create systemic bias against disfluent speech \cite{mujtaba-etal-2024-lost}.

While a lot of research has been published on disfluent speech detection and classification, a less-explored problem for stuttered speech research is \textbf{evaluation reference}; 
what should the output of ASR systems look like for stuttered speech? The \emph{answer}, we argue, cannot be decoupled from the target application. For a 2-second speech recording that includes multiple stuttering events, ASR model \texttt{\textbf{X}} produces the transcription \emph{`and I got it'}, while ASR model \texttt{\textbf{Y}} produces the transcription \emph{`and i got i i got it'}. Which of these should be considered more accurate depends on the downstream task and its intended usage.

ASR evaluation typically involves the calculation of Word Error Rate (WER), which presumes the existence of a reference transcription.  In atypical speech research, the most common practice is to use the intended transcript as  reference \cite{mujtaba-etal-2024-lost}, and  verbatim transcripts are often formatted post-hoc to remove disfluency events \cite{Xue2024FindingsOT} for use as reference, which in this example would rank ASR model \texttt{\textbf{X}} as better than ASR model \texttt{\textbf{Y}}. We argue this ranking is overly \emph{reductive} stripping away complexities associated with use-case and context common to atypical speech. For example, for  a dictation model, a PWS dictating an article would prefer \texttt{\textbf{X}} over \texttt{\textbf{Y}}, while for speech therapy and diagnostic purposes, a therapist would prefer \texttt{\textbf{Y}} to \texttt{\textbf{X}} for identifying characters/words that induce stuttering events. We solidify this distinction between ground truth references for atypical ASR and define the two forms below:
\begin{itemize}
    \item \textbf{Intended transcription} captures the intended message with disfluencies removed \emph{(e.g., `I want to go')}. Typical use cases for this includes dictation, command/control. Voice-assistant tools rely on intended transcription for accurate command execution \cite{Mujtaba2025FineTuningAF}.
    \item \textbf{Verbatim transcription} captures what was said, including disfluencies \emph{(e.g., `I I I want to uhmm go')}. Typical use cases for this includes clinical assessment, research on stuttering patterns, and contexts where preserving one’s produced speech matters. Clinician workflows often rely on verbatim transcripts with disfluency codes, but this is time-consuming and tools are limited \cite{Heeman2016UsingCA}.

\end{itemize}

Our review of recent\footnote{2020 - 2025} ASR works published in main speech conferences\footnote{Interspeech, ICASSP, ASRU, IEEE SLT} show an implicit dominance of intended ASR; \textbf{only one} work \cite{Sridhar2025JjjjustSB} \textit{differentiates} \textit{and evaluates} both intended and verbatim speech. Some approaches explicitly apply post-hoc “disfluency refinements” \cite{Xue2024FindingsOT} (e.g., removing filler words), which may improve “intended-like” metrics while harming verbatim fidelity without explicitly clarifying the intended use-case of the model. Mainstream atypical ASR research typically reports a single WER against a single reference, often without specifying the reference type, implicitly treating an arbitrary transcript style as universally correct. 
\vspace{0.5em}

\noindent \textbf{Why Does it Matter?} Atypical ASR supports a wider set of stakeholders than just PWS, including speech-language pathologists (SLPs) and parents/guardians who may need accurate analyses of what was said for tasks such as severity assessment or identifying problematic words. Current mainstream practices raises a fairness and accessibility issue: systems reported as \textit{``best”} for atypical speech can be misleading if they silently normalize or remove stuttering attributes without explicitly stating that they are optimized for semantic transcription (intended) rather than verbatim reporting. Because the same utterance can have multiple \textit{valid} textual references (intended and verbatim), choosing one without disclosure effectively hard-codes a normative decision about how a person's speech should be represented.

\vspace{0.5em}
\noindent In this work: (i) We evaluate 11 ASR models on FluencyBank Timestamped~\cite{Romana2024FluencyBankTA} using \textbf{both} transcript forms as references, quantifying how different architecture families behave under each `\emph{ground truth}' and showing why single-reference evaluation is insufficient. (ii) We additionally align the audio segments in FluencyBank Timestamped with the recently proposed clinical stutter events annotations (CASA) \cite{valente25_interspeech} for FluencyBank to examine the impact of stuttering events on the models' performance in both cases. (iii) We finally make recommendations on architectures to use for specific use-case, discuss the potential harms of overlooking this dual reference problem and highlight best practices for building inclusive speech technology for all voices and speech types.


\section{Related Work}

\noindent \textbf{Performance and Biases of ASR for Atypical Speech.} Mujtaba et al. \cite{mujtaba-etal-2024-lost} evaluated six ASR systems on real and synthetic stuttered speech using intended transcripts as the sole reference, quantifying ASR bias by comparing performance on fluent versus stuttered speech with both WER and BERTScore.   Notably, some works explicitly optimize for intended transcription by treating ASR as a semantic problem \cite{Mitra2021AnalysisAT, Mujtaba2025FineTuningAF} for voice commands, whereas many studies implicitly adopt intended transcripts as the reference without stating the assumption. Sridhar and Wu \cite{Sridhar2025JjjjustSB} benchmarked Whisper's performance on stuttered speech and were the first to evaluate ASR model's output against both intended and verbatim references. However the study is limited to the Whisper family (v2/v3). Our work standardizes and extends this line of inquiry by formalizing \textbf{verbatim} vs. \textbf{intended} transcripts as two valid references for atypical speech. We benchmark which modelling paradigm best supports each use case across a broader set of models and discuss how mainstream evaluation metrics perform in intended speech recognition settings.


\vspace{0.5em}

\noindent \textbf{ASR for Disfluent Speech Identification.} Several works have proposed identifying disfluent events in speech through transcription; producing \textit{pseudo-verbatim} transcripts with special disfluency tokens \cite{Zhang2025AnalysisAE, UDM, Alharbi2017AutomaticRO} \textit{(e.g. I$\backslash$r won \textbf{or} I $\prec$rep$\succ$ won \textbf{for} I I I won)}. In \cite{Mendelev2020ImprovedRT}, the authors proposed labeling transcripts with special stutter event tokens, disfluent data augmentation and model fine-tuning to improve WER. This approach treats ASR as disfluent event transcription where disfluent events are marked with special tokens in produced transcription \textit{(e.g. I \textbf{$\prec$rep$\succ$} won vs. I \textbf{I I} won, \textbf{$\prec$rep$\succ$} indicates repetition event in this example}). Jiachen et al \cite{UDM} introduced \textit{imperfect WER} to measure intended ASR performance for aphasia speech which is needed in their pipeline for disfluent event detection.
 

\section{Methodology}

\noindent \textbf{Benchmarked Models.} Our benchmark covers various ASR modeling paradigms. We compare these paradigms under the hypothesis that architectural and decoding design choices can systematically favor one transcription reference over the other: verbatim  versus intended, independent of dataset-specific factors. Although not all benchmarked models (shown in Table \ref{tab:models_tested}) have similar data settings, the NVIDIA-family of models are mostly trained on their pre-defined NeMo ASRSET\footnote{\url{https://huggingface.co/nvidia/stt_en_conformer_ctc_large\#datasets}}, this allows us to directly compare models trained with different training configurations on the same data. \textbf{(i) Autoregressive ASR models:} These models typically transcribe a sequence of texts using all audio information (global context) and previously predicted text tokens. This approach is similar to language modelling in that current predictions are conditioned on previously produced text. We hypothesize models that are trained in this way are more likely to be better at \textbf{intended} speech recognition since the decoder can use preceding text context to \textit{predict} regions of uncertain acoustics (e.g prolongations, repetitions). \textbf{Transducer models} are streaming models \cite{gulati20_interspeech} that predict the next text token using only past acoustic frames up to the current time step together with the predicted text token history. \textbf{(ii) CTC ASR models:} Models trained with the CTC loss \cite{CTC} only use the encoder’s frame-level representations of acoustic information at current time step to generate probability distribution over output text token units. They typically assumes that each output unit is conditionally independent of others. As a result, they rely less on token history and more directly on the local acoustic evidence. We hypothesize these models will be better for \textbf{verbatim transcription} since the models seem to map current acoustics to text token.

\begin{table}[h]
    \centering
    \footnotesize
    \caption{Models benchmarked in this paper.}
    \label{tab:models_tested}
    \begin{tabular}{@{}llc@{}} 
        \toprule
        \textbf{Configuration} & \textbf{Model} & \textbf{Ref.} \\ 
        \midrule
        \multicolumn{3}{@{}l}{\textit{Autoregressive}} \\
        \quad Seq2Seq & Whisper Large-v3 & \cite{radford2022robustspeechrecognitionlargescale} \\
        \quad Seq2Seq & SpeechBrain Transformer & \cite{speechbrain} \\
        \quad Seq2Seq & NVIDIA Canary-1B-v2 & \cite{sekoyan2025canary1bv2parakeettdt06bv3efficient} \\
        \quad Conformer-Transducer & NVIDIA Transducer & \cite{gulati20_interspeech} \\
        \multicolumn{3}{@{}l}{\textit{Non-Autoregressive}} \\
        \quad Conformer CTC & NVIDIA CTC & \cite{speechbrain} \\
        \quad Encoder CTC & HuBERT Large & \cite{hubert} \\
        \quad Encoder CTC & Wav2Vec2 Large & \cite{wav2vec} \\
        \quad Conformer CTC & NVIDIA Fast Conformer & \cite{fastconformer} \\
        \quad Conformer CTC & SpeechBrain Streaming & \cite{speechbrain} \\
        \quad Convolutional Encoder CTC & NVIDIA QuartzNet & \cite{Kriman2019QuartznetDA} \\
        \quad CTC + Attention & SpeechBrain CRDNN & \cite{speechbrain} \\ 
        \bottomrule
    \end{tabular}
\end{table}


\noindent \textbf{Datasets:} FluencyBank Timestamped \cite{Romana2024FluencyBankTA} and CASA \cite{valente25_interspeech} are the two main sources utilized for their transcriptions and stutter event labels respectively. FluencyBank Timestamped contains \textbf{verbatim} and \textbf{intended} transcriptions for the interview section of FluencyBank \cite{FluencyBank}. CASA introduces clinical segmentation of primary and secondary stuttering events to the FluencyBank corpus. We align (i.e match overlapping audio event to transcript segments) the audio segments from FluencyBank Timestamped to \emph{any} event labels in CASA to further examine the effect of stuttering events on \textbf{intended} and \textbf{verbatim} transcription models. The dataset is in English and we used the full set (containing $3430$ samples) in FluencyBank Timestamped for this work. The primary stuttering events correlating to transcription from CASA are: (1) \textbf{SR}: Syllable Repetition, repeated movement of an entire syllable; (2) \textbf{ISR}: Incomplete Syllable Repetition, repetition of parts of syllables; (3) \textbf{MUR}: Multisyllable Unit Repetition, repetitions involving multiple syllables; (4) \textbf{P}: sound Prolongation; and (5) \textbf{B}: Blocks \cite{valente25_interspeech}.

\vspace{0.5em}

\noindent \textbf{Inference.} \emph{Model configuration.} We chose open-source models and utilized the publicly available default configuration for inference for each model to ensure reproducibility and avoid bias. We evaluate all models \emph{as-is} without any fine-tuning or parameter-tuning. We use the inference configuration, decoding settings and setup described on the Hugging face\footnote{\url{https://huggingface.co/models}} model page for all models. Code available here.\footnote{\url{https://github.com/Theehawau/usecase_asr}}

\noindent \textbf{Evaluation metrics:} We primarily evaluated the models using  \textit{word error rate} (WER) obtaining intended speech WER (isWER) and verbatim WER (vWER) when comparing against intended and verbatim reference respectively. We also report SeMaScore~\cite{sasindran24_interspeech}, a recently introduced metric reported to outperform BERTScore on correlation to human assessment and highlighted to have strong capabilities on atypical speech for the intended speech recognition task. We report model rank \emph{x-rank} based on WER to compare models behaviour on the two references. To standardize evaluation, we adjust casing and punctuation using the
\textit{BasicTextNormalizer}\footnote{\url{https://huggingface.co/docs/transformers/en/model_doc/whisper\#transformers.WhisperTokenizer.basic_normalize}} function from Whisper on predicted texts and references before evaluation. This function removes punctuations and standardizes casing \emph{without} altering repeated tokens or removing fillers.

\section{Results}

\subsection{ASR Model Ranking}
\label{sec:rankModels}
 
 As we hypothesized, when we compare ASR models predictions against the two references we see inconsistent performance rankings. NVIDIA CTC model Ranks first for providing verbatim transcriptions but ranks 5th for intended transcription. Interestingly, whisper\cite{radford2022robustspeechrecognitionlargescale}, which is often benchmarked for atypical speech recognition \cite{Sridhar2025JjjjustSB, Mujtaba2025FineTuningAF}, ranks second for intended speech and third for verbatim (see Table \ref{tab:modelrank}). 
 
\begin{figure}[h]
    \centering
    \includegraphics[width=1\linewidth]{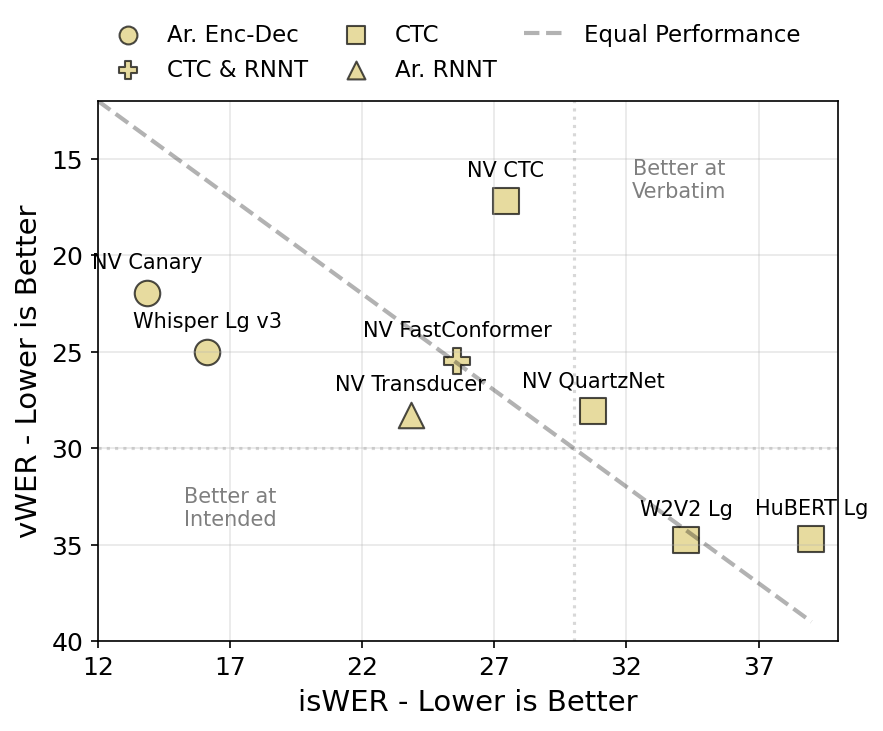}
    \caption{Intended vs Verbatim ASR Performance for top performing models. Distance from diagonal infers task specialization. Ar: Autoregressive, Enc-Dec: Encoder-Decoder. }
    \label{fig:resultArchitecture}
    \vspace{-1.5em}
\end{figure}

 Figure \ref{fig:resultArchitecture} shows the \textit{specialization} of some of the high performing models we tested. We can clearly see a pattern in model training configuration and model speciality. The autoregressive sequence-to-sequence style models are found to be better at intended speech recognition while models trained with CTC loss perform better for verbatim speech recognition.  While the models in Figure \ref{fig:resultArchitecture} are not entirely comparable in terms of training data and number of parameters, the NVIDIA-family models are trained on the same Nemo ASRSET: the training data is the same for these model variants and their performance follow the same general pattern, providing evidence that the training paradigm alone can bias the type of transcription the model produces, regardless of the type of transcription used for training.


 \begin{table}[h]
     \caption{Verbatim and Intended results for all models. \colorbox[HTML]{38FFF8}{Best model for Intended ASR}, \colorbox[HTML]{6665CD}{Best model for Verbatim ASR}. N = NVIDIA; SB = SpeechBrain.}
     \centering
     \footnotesize
     \resizebox{\columnwidth}{!}{%
     \begin{tabular}{@{}lccccc@{}} \toprule
         \textbf{Model} & \textbf{isWER} ($\downarrow$) & \textbf{SeMaScore} ($\uparrow$) & \textbf{isRank} & \textbf{vWER} ($\downarrow$)  & \textbf{vRank} \\ \midrule
         \multicolumn{6}{@{}l}{\textit{Autoregressive}} \\
         \colorbox[HTML]{38FFF8}{N Canary-1B-v2} & \textbf{13.85} & 0.91 & 1 & 21.95 & 2 \\
Whisper Large v3 & 16.13 & 0.92 & 2 & 25.01 & 3 \\
SB Transformer & 72.60 & 0.63 & 10 & 62.13 & 10 \\
N Transducer & 23.84 & 0.83 & 3 & 28.30 & 6 \\ \cmidrule{1-6}
         \multicolumn{6}{@{}l}{\textit{Non-Autoregressive}} \\
\colorbox[HTML]{6665CD}{N CTC} & 27.43 & 0.83 & 5 & \textbf{17.20} & 1 \\
N FastConformer & 25.60 & 0.83 & 4 & 25.49 & 4 \\
SB CRDNN & 72.76 & 0.57 & 11 & 66.04 & 11 \\
Wav2Vec2 Large & 34.26 & 0.75 & 7 & 34.75 & 8 \\
HuBERT Large & 39.00 & 0.74 & 8 & 34.73 & 7 \\
N QuartzNet & 30.75 & 0.80 & 6 & 28.10 & 5 \\
SB Streaming & 47.08 & 0.62 & 9 & 42.81 & 9 \\ \bottomrule
     \end{tabular}
     }
     \label{tab:modelrank}
     \vspace{-1.5em}
 \end{table}



\subsection{ASR in the Presence of Atypical Speech Events} 
All \textbf{verbatim} results discussed in this section are obtained from the predictions of the NVIDIA CTC model compared with the verbatim transcript as reference. All \textbf{intended} result discussed in this section are obtained from the predictions of the NVIDIA canary-1B-v2 model compared with the intended transcript as reference. These models rank 1st for each use-case from the experiments in Section \ref{sec:rankModels}.

\begin{figure}[h]
    \centering
    \includegraphics[width=1\linewidth]{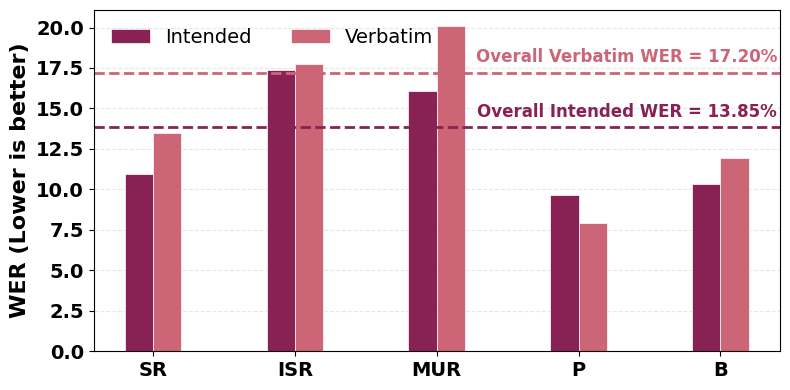}
\caption{Word error rate by stutter event type or \textbf{intended} and \textbf{verbatim} references.}
    \label{fig:werByEvent}
    \vspace{-0.5em}
\end{figure}

\noindent \textbf{WER and stutter speech events.} We show the effect of different stutter events on ASR metric in Figure \ref{fig:werByEvent}. We find that prolongation events (\texttt{P}) are easiest (lowest WER) for models to handle in both use case, since the intended lexicon is usually stable and only the acoustic counterpart is time-stretched but still consistent enough for the model to map to the correct word. Multisyllable unit repetition (\texttt{MUR}) and  incomplete syllable repetition (\texttt{ISR}) are hardest for the models for both use cases.  For intended use case, the model struggles to get rid of disfluent fragments/repetitions. For verbatim use case, the models struggle to preserve the repeated word correctly (what fragment was produced). We show an example prediction in Table \ref{tab:multipleEvents} to illustrate this. 

\begin{table}[ht!]
    \centering
    \footnotesize
    \caption{Prediction output for an audio sample containing \texttt{SR}, \texttt{MUR} and \texttt{B} events.}
    \begin{tabular}{ll} \toprule
         \textbf{Model}& \textbf{Prediction}\\ \midrule
         V Ground Truth & i had the th the th th therapy \\
         N CTC & i had the ta the ta they they a rapi \\ \midrule
         I Ground Truth & i had therapy \\
         N Canary  & I had the tape, the tape, the tape, the repeat. \\ \bottomrule
    \end{tabular}
    \label{tab:multipleEvents}
\end{table}


\begin{table}[h]
    \centering
    \caption{Some sample model outputs with errors. V = Verbatim, I = Intended, model $\#$1 = best model by WER.}
    \resizebox{\columnwidth}{!}{%
    \begin{tabular}{@{}l|l|l@{}} \toprule
         \textbf{Event} & \textbf{Model} & \textbf{Prediction} \\ \midrule
      \multirow{2}{*}{\texttt{ISR}} & V Ground Truth & just kind of \underline{bur} bureaucratic stuff \\
         & V model $\#$1 & just kind of \textbf{bu} bureaucratic stuff \\ \cmidrule(rr){1-1} \cmidrule(lr){2-2} \cmidrule(ll){3-3} 
         \multirow{2}{*}{\texttt{ISR}} & I Ground Truth & just kind of bureaucratic stuff \\
         & I model $\#$1 & just kind of \textbf{your} bureaucratic \textbf{self} \\ 
         \midrule
         \multirow{2}{*}{\texttt{MUR}} & V Ground Truth & \underline{you're} \underline{you're} \underline{you're} a \underline{bully} \\
         & V model $\#$1 & \textbf{you} \textbf{your} \textbf{your} \textbf{bolly} \\ \cmidrule(rr){1-1} \cmidrule(lr){2-2} \cmidrule(ll){3-3} 
         \multirow{2}{*}{\texttt{B}} & I Ground Truth & for \underline{uh as long as i was in} the workforce \\
         & I model $\#$1 & for \textbf{long-standing} the workforce \\ \bottomrule
    \end{tabular}
    }
    \label{tab:sampleOutputs}
    \vspace{-1.5em}
\end{table}


\noindent \textbf{What drives use-case WER?} We examined the model outputs a little deeper for both use-case specifically for the speech of PWS that exclusively contains certain stuttering events. We examine the composition of edit errors: Substituion, Insertion and Deletion, relative to each valid transcript. We find that \textit{insertions} and \textit{substitutions} (Sub=115; Del=53; Ins=112) mostly dominate for \textbf{intended} speech recognition while \textit{substitutions} (Sub=199; Del=94; Ins=74) mostly dominate for \textbf{verbatim} speech recognition. In Table \ref{tab:sampleOutputs}, the \textbf{verbatim} transcription for the \texttt{ISR} event, shows a substitution (\textit{bu vs bur}) driven by the models inability to model correctly the incomplete syllable. For the same audio sample, we see the intended model tries to match the acoustic event to a plausible transcription (\textit{inserting your and substituting stuff with self}).

\subsection{Discussion}

\textbf{Dual references change what it means to be “accurate”.}
Our results show that model ranking is not stable across different evaluation references: the top model using intended reference (NVIDIA Canary-1B-v2, isWER $13.85$) is not the top model using verbatim reference (NVIDIA CTC, vWER $17.2$), and several systems shift substantially in rank when switching references (Table~\ref{tab:modelrank}). This instability is not a minor scoring artifact; it reflects that atypical speech can legitimately map to multiple textual outputs depending on downstream goals, and evaluation cannot be decoupled from use cases. 

\vspace{0.5em}

\noindent \textbf{Architectural specialization suggests predictable trade-offs.} Across model families, we observe a consistent pattern: autoregressive sequence-to-sequence systems (e.g., Canary, Whisper) tend to perform better on intended transcription, while CTC-based systems tend to perform better on verbatim transcription (Figure~\ref{fig:resultArchitecture}). This aligns with the intuition that autoregressive decoders can use linguistic context to \emph{normalize} acoustically uncertain regions (e.g., repetitions or fragments), whereas CTC-style models rely more heavily on local acoustic evidence and thus more readily preserves disfluencies. Importantly, within the subset of NVIDIA models trained on the same data source (NeMo ASRSET), the same specialization trend holds, suggesting this effect is not solely explained by dataset differences but also by training objectives and decoding behavior.

\vspace{0.5em}

\noindent \textbf{Measuring intended accuracy.} Intended SR evaluation involves more nuances than typical ASR, giving the reference might not exactly map to the acoustics by simply removing duplicates, and conveying \emph{intent} is important. We examine isWER and popular semantic metrics: BERTScore\cite{BERTScore} and SeMaScore\cite{sasindran24_interspeech}. In Table~\ref{tab:ASR_METRICS}, the three metrics behave differently enough that \emph{none is sufficient on its own} for reporting intended ASR accuracy. isWER is the most transparent and widely used, but it is also the most surface-form strict: Whisper’s “\emph{20 percent environmental}” is penalized (isWER=0.33) against “\emph{twenty percent environmental}” even though the intended meaning is unchanged, showing that isWER can overstate errors for acceptable formatting variants (numbers, tokenization) and cannot indicate whether a mistake is meaning-preserving or meaning-breaking. Semantic metrics help with that, but they also have blind spots: SeMaScore separates meaning-changing confusions more clearly (\emph{“bully” vs “bullet”} in \texttt{V GT 2} yields a much lower SeMaScore for the worse hypothesis), yet it can still assign a relatively low score to meaning-equivalent variants (\emph{“20” vs “twenty”} in \texttt{V GT 1}), suggesting sensitivity to representation choices unless normalization is carefully controlled. BERTScore is even less discriminative in these short utterances: it stays high for clearly wrong hypotheses (\emph{“your your your bullet”} still gets a high similarity score of 0.84) and can fail to separate distinct named-entity errors (both \emph{“to calgary center”} and \emph{“to cali or centers”} receive BERT=0.86 despite very different isWER).








\begin{table}[t]
\caption{WER, SeMaScore and BERTScore for sample atypical speech transcripts for \textbf{intended} ASR. V GT = Verbatim Ground Truth, I GT = Intended Ground Truth, I $\# 1$ = NVIDIA Canary-1B-v2 , I $\# 2$ = Whisper largev3}
\resizebox{\columnwidth}{!}{%
\begin{tabular}{@{}l p{0.58\columnwidth} p{0.10\columnwidth} p{0.10\columnwidth} p{0.12\columnwidth}@{}}
\toprule
\textbf{Model} & \textbf{Transcript} & \textbf{isWER}$\downarrow$ & \textbf{SeMa.}$\uparrow$ & \textbf{BERT.}$\uparrow$ \\ \midrule
V GT 1 & tw twenty percent environmental &  &  &  \\ 
I GT & twenty percent environmental & 0 & 1 & 1 \\
I $\# 1$ & twenty percent environmental & 0 & 1 & 1 \\ 
I $\# 2$ & 20 percent environmental & 0.33 & 0.63 & 0.99 \\
\midrule
V GT 2 & you're you're you're a bully &  &  &  \\ 
I GT & you're a bully & 0 & 1 & 1 \\
I $\# 1$ & you're bullied & 0.5 & 0.61 & 0.94 \\
I $\# 2$  & your your your bullet & 1 & 0.37 & 0.84 \\ \midrule
V GT 3 & to callier center &  &  &  \\
I GT & to callier center & 0 & 1 & 1 \\
I $\# 1$  & to calgary center & 0.3 & 0.7 & 0.86 \\ 
I $\# 2$  & to cali or centers & 1 & 0.64 & 0.86 \\
\bottomrule
\end{tabular}%
}
\label{tab:ASR_METRICS}
\vspace{-1.5em}

\end{table}
\section{Conclusion}
We argued that atypical ASR involves two valid transcription references \emph{intended} and \emph{verbatim} and showed that conflating them into a single reference can misrepresent model capability. Benchmarking 11 ASR systems on FluencyBank Timestamped with both references, we found substantial rank instability: autoregressive seq2seq models tend to specialize in intended transcription, while CTC-based models tend to specialize in verbatim transcription, implying that “best” is use-case dependent rather than universal. By aligning FluencyBank segments with CASA stutter-event labels, we further identified how event types such as ISR and MUR disproportionately drive errors, and how dominant edit types differ by reference, revealing distinct failure modes for intended vs.\ verbatim ASR. We recommend atypical speech recognition papers should explicitly state what form of transcription they are optimizing for and be open about possible limitations of their model (like retention of repetitions). For intended-facing applications (e.g., dictation), evaluations should use intended references and report isWER together with a semantic similarity metric. For verbatim-facing applications (e.g., clinical assessment), evaluations should use verbatim references and report vWER. This work is limited to English and to the FluencyBank Timestamped domain; broader conclusions will benefit from multilingual datasets and additional speaking contexts.

\section{Generative AI Use Disclosure}

Generative AI tool (Writefull) use is limited to sentence polishing for clarity of presentation. All experimentation and interpretations were carried out solely by the authors.

\bibliographystyle{IEEEtran}
\bibliography{mybib}

@article{Sridhar2025JjjjustSB,
  title={J-j-j-just Stutter: Benchmarking Whisper's Performance Disparities on Different Stuttering Patterns},
  author={Charan Sridhar and Shaomei Wu},
  journal={Interspeech},
  year={2025},  
  url={https://api.semanticscholar.org/CorpusID:281308675}
}

@inproceedings{Zhang2025AnalysisAE,
  title     = {{Analysis and Evaluation of Synthetic Data Generation in Speech Dysfluency Detection}},
  author    = {Jinming Zhang and Xuanru Zhou and Jiachen Lian and Shuhe Li and William Li and Zoe Ezzes and Rian Bogley and Lisa Wauters and Zachary Miller and Jet Vonk and Brittany Morin and Maria Gorno-Tempini and Gopala Anumanchipalli},
  year      = {2025},
  booktitle = {{Interspeech}},
  pages     = {1853--1857},
  doi       = {10.21437/Interspeech.2025-2658},
  issn      = {2958-1796},
}

@inproceedings{Mitra2021AnalysisAT,
  title={Analysis and Tuning of a Voice Assistant System for Dysfluent Speech},
  author={Vikramjit Mitra and Zifang Huang and Colin S. Lea and Lauren Tooley and Sarah Wu and Darren Botten and Ashwin Palekar and Shrinath Thelapurath and Panayiotis G. Georgiou and Sachin S. Kajarekar and Jefferey Bigham},
  booktitle={Interspeech},
  year={2021},
  url={https://api.semanticscholar.org/CorpusID:235593228}
}

@article{Romana2024FluencyBankTA,
  title={FluencyBank Timestamped: An Updated Data Set for Disfluency Detection and Automatic Intended Speech Recognition},
  author={Amrit Romana and Minxue Niu and M. Perez and Emily Mower Provost},
  journal={Journal of Speech, Language, and Hearing Research : JSLHR},
  year={2024},
  volume={67},
  pages={4203 - 4215},
  url={https://api.semanticscholar.org/CorpusID:273200647}
}

@inproceedings{mujtaba-etal-2024-lost,
    title = "Lost in Transcription: Identifying and Quantifying the Accuracy Biases of Automatic Speech Recognition Systems Against Disfluent Speech",
    author = "Mujtaba, Dena  and
      Mahapatra, Nihar  and
      Arney, Megan  and
      Yaruss, J  and
      Gerlach-Houck, Hope  and
      Herring, Caryn  and
      Bin, Jia",
    editor = "Duh, Kevin  and
      Gomez, Helena  and
      Bethard, Steven",
    booktitle = "Proceedings of the 2024 Conference of the North American Chapter of the Association for Computational Linguistics: Human Language Technologies (Volume 1: Long Papers)",
    month = jun,
    year = "2024",
    address = "Mexico City, Mexico",
    publisher = "Association for Computational Linguistics",
    url = "https://aclanthology.org/2024.naacl-long.269/",
    doi = "10.18653/v1/2024.naacl-long.269",
    pages = "4795--4809"
}

@inproceedings{user-perceptions-technical,
title = {From User Perceptions to Technical Improvement: Enabling People Who Stutter to Better Use Speech Recognition},
booktitle = {CHI},
author = {Colin Lea and Zifang Huang and Jaya Narain and Lauren Tooley and Dianna Yee and Tien Dung Tran and Panayiotis Georgiou and Jeffrey Bigham and Leah Findlater},
year = {2023},
note = {Available at: \url{https://arxiv.org/abs/2302.09044}}
}

@inproceedings{Heeman2016UsingCA,
  title={Using Clinician Annotations to Improve Automatic Speech Recognition of Stuttered Speech},
  author={Peter A. Heeman and Rebecca Lunsford and Andy McMillin and J Scott Yaruss},
  booktitle={Interspeech},
  year={2016},
  url={https://api.semanticscholar.org/CorpusID:1906213}
}

@article{FluencyBank,
author = {Ratner, Nan and Macwhinney, Brian},
year = {2018},
month = {03},
pages = {},
title = {Fluency Bank: A new resource for fluency research and practice},
volume = {56},
journal = {Journal of Fluency Disorders},
doi = {10.1016/j.jfludis.2018.03.002}
}

@inproceedings{Mujtaba2025FineTuningAF,
  title     = {{Fine-Tuning ASR for Stuttered Speech: Personalized vs. Generalized Approaches}},
  author    = {Dena Mujtaba and Nihar R. Mahapatra},
  year      = {2025},
  booktitle = {{Interspeech 2025}},
  pages     = {3568--3572},
  doi       = {10.21437/Interspeech.2025-2373},
  issn      = {2958-1796},
}

@inproceedings{radford2022robustspeechrecognitionlargescale,
author = {Radford, Alec and Kim, Jong Wook and Xu, Tao and Brockman, Greg and McLeavey, Christine and Sutskever, Ilya},
title = {Robust speech recognition via large-scale weak supervision},
year = {2023},
publisher = {JMLR.org},
booktitle = {Proceedings of the 40th International Conference on Machine Learning},
articleno = {1182},
numpages = {27},
location = {Honolulu, Hawaii, USA},
series = {ICML'23}
}

@article{Mendelev2020ImprovedRT,
  title={Improved Robustness to Disfluencies in Rnn-Transducer Based Speech Recognition},
  author={Valentin Mendelev and Tina Raissi and Guglielmo Camporese and Manuel Giollo},
  journal={IEEE International Conference on Acoustics, Speech and Signal Processing (ICASSP)},
  year={2021},
  pages={6878-6882},
  url={https://api.semanticscholar.org/CorpusID:228376178}
}

@inproceedings{Alharbi2017AutomaticRO,
  title={Automatic recognition of children's read speech for stuttering application},
  author={Sadeen Alharbi and Anthony J. H. Simons and Shelagh Brumfitt and Phil D. Green},
  booktitle={Workshop on Child, Computer and Interaction},
  year={2017},
  url={https://api.semanticscholar.org/CorpusID:14030764}
}

@inproceedings{valente25_interspeech,
      title = {{Clinical Annotations for Automatic Stuttering Severity Assessment}},
      author = {Ana Valente and Rufael Marew and Hawau Toyin and Hamdan Al-Ali and Anelise Bohnen and Inma Becerra and Elsa Soares and Gonçalo Leal and Hanan Aldarmaki},
      year = {2025},
      booktitle = {{Interspeech 2025}},
      pages = {4318--4322},
      doi = {10.21437/Interspeech.2025-1916},
      issn = {2958-1796},
      }

@article{Xue2024FindingsOT,
  title={Findings of the 2024 Mandarin Stuttering Event Detection and Automatic Speech Recognition Challenge},
  author={Hongfei Xue and Rong Gong and Mingchen Shao and Xin Xu and Lezhi Wang and Lei Xie and Hui Bu and Jiaming Zhou and Yong Qin and Jun Du and Ming Li and Binbin Zhang and Bin Jia},
  journal={2024 IEEE Spoken Language Technology Workshop (SLT)},
  year={2024},
  pages={385-392},
  url={https://api.semanticscholar.org/CorpusID:272524924}
}

@inproceedings{CTC,
author = {Graves, Alex and Fern\'{a}ndez, Santiago and Gomez, Faustino and Schmidhuber, J\"{u}rgen},
title = {Connectionist temporal classification: labelling unsegmented sequence data with recurrent neural networks},
year = {2006},
isbn = {1595933832},
publisher = {Association for Computing Machinery},
address = {New York, NY, USA},
url = {https://doi.org/10.1145/1143844.1143891},
doi = {10.1145/1143844.1143891},
booktitle = {Proceedings of the 23rd International Conference on Machine Learning},
pages = {369–376},
numpages = {8},
location = {Pittsburgh, Pennsylvania, USA},
series = {ICML '06}
}

@inproceedings{wav2vec,
author = {Baevski, Alexei and Zhou, Henry and Mohamed, Abdelrahman and Auli, Michael},
title = {wav2vec 2.0: a framework for self-supervised learning of speech representations},
year = {2020},
isbn = {9781713829546},
publisher = {Curran Associates Inc.},
address = {Red Hook, NY, USA},
booktitle = {Proceedings of the 34th International Conference on Neural Information Processing Systems},
articleno = {1044},
numpages = {12},
location = {Vancouver, BC, Canada},
series = {NIPS '20}
}

@misc{speechbrain,
  title={{SpeechBrain}: A General-Purpose Speech Toolkit},
  author={Mirco Ravanelli and Titouan Parcollet and Peter Plantinga and Aku Rouhe and Samuele Cornell and Loren Lugosch and Cem Subakan and Nauman Dawalatabad and Abdelwahab Heba and Jianyuan Zhong and Ju-Chieh Chou and Sung-Lin Yeh and Szu-Wei Fu and Chien-Feng Liao and Elena Rastorgueva and François Grondin and William Aris and Hwidong Na and Yan Gao and Renato De Mori and Yoshua Bengio},
  year={2021},
  eprint={2106.04624},
  archivePrefix={arXiv},
  primaryClass={eess.AS},
  note={arXiv:2106.04624}
}

@misc{sekoyan2025canary1bv2parakeettdt06bv3efficient,
      title={Canary-1B-v2 \& Parakeet-TDT-0.6B-v3: Efficient and High-Performance Models for Multilingual ASR and AST}, 
      author={Monica Sekoyan and Nithin Rao Koluguri and Nune Tadevosyan and Piotr Zelasko and Travis Bartley and Nikolay Karpov and Jagadeesh Balam and Boris Ginsburg},
      year={2025},
      eprint={2509.14128},
      archivePrefix={arXiv},
      primaryClass={cs.CL},
      url={https://arxiv.org/abs/2509.14128}, 
}

@inproceedings{fastconformer,
  author       = {Dima Rekesh and
                  Nithin Rao Koluguri and
                  Samuel Kriman and
                  Somshubra Majumdar and
                  Vahid Noroozi and
                  He Huang and
                  Oleksii Hrinchuk and
                  Krishna C. Puvvada and
                  Ankur Kumar and
                  Jagadeesh Balam and
                  Boris Ginsburg},
  title        = {Fast Conformer With Linearly Scalable Attention For Efficient Speech
                  Recognition},
  booktitle    = {{IEEE} Automatic Speech Recognition and Understanding Workshop, {ASRU}
                  2023, Taipei, Taiwan, December 16-20, 2023},
  pages        = {1--8},
  publisher    = {{IEEE}},
  year         = {2023},
  url          = {https://doi.org/10.1109/ASRU57964.2023.10389701},
  doi          = {10.1109/ASRU57964.2023.10389701},
  bibsource    = {dblp computer science bibliography, https://dblp.org}
}

@inproceedings{gulati20_interspeech,
  title     = {{Conformer: Convolution-augmented Transformer for Speech Recognition}},
  author    = {Anmol Gulati and James Qin and Chung-Cheng Chiu and Niki Parmar and Yu Zhang and Jiahui Yu and Wei Han and Shibo Wang and Zhengdong Zhang and Yonghui Wu and Ruoming Pang},
  year      = {2020},
  booktitle = {{Interspeech 2020}},
  pages     = {5036--5040},
  doi       = {10.21437/Interspeech.2020-3015},
  issn      = {2958-1796},
}

@article{Kriman2019QuartznetDA,
  title={Quartznet: Deep Automatic Speech Recognition with 1D Time-Channel Separable Convolutions},
  author={Samuel Kriman and Stanislav Beliaev and Boris Ginsburg and Jocelyn Huang and Oleksii Kuchaiev and Vitaly Lavrukhin and Ryan Leary and Jason Li and Yang Zhang},
  journal={IEEE International Conference on Acoustics, Speech and Signal Processing (ICASSP)},
  year={2020},
  pages={6124-6128},
}

@article{hubert,
author = {Hsu, Wei-Ning and Bolte, Benjamin and Tsai, Yao-Hung Hubert and Lakhotia, Kushal and Salakhutdinov, Ruslan and Mohamed, Abdelrahman},
title = {HuBERT: Self-Supervised Speech Representation Learning by Masked Prediction of Hidden Units},
year = {2021},
issue_date = {2021},
publisher = {IEEE Press},
volume = {29},
issn = {2329-9290},
url = {https://doi.org/10.1109/TASLP.2021.3122291},
doi = {10.1109/TASLP.2021.3122291},
journal = {IEEE/ACM Trans. Audio, Speech and Lang. Proc.},
month = oct,
pages = {3451–3460},
numpages = {10}
}

@article{UDM,
  title={Unconstrained Dysfluency Modeling for Dysfluent Speech Transcription and Detection},
  author={Jiachen Lian and Carly Feng and Naasir Farooqi and Steve Li and Anshul Kashyap and Cheol Jun Cho and Peter Wu and Robin Netzorg and Tingle Li and Gopala Krishna Anumanchipalli},
  journal={2023 IEEE Automatic Speech Recognition and Understanding Workshop (ASRU)},
  year={2023},
  pages={1-8},
  url={https://api.semanticscholar.org/CorpusID:266374895}
}

@inproceedings{sasindran24_interspeech,
  title     = {{SeMaScore: A new evaluation metric for automatic speech recognition tasks}},
  author    = {Zitha Sasindran and Harsha Yelchuri and T. V. Prabhakar},
  year      = {2024},
  booktitle = {{Interspeech}},
  pages     = {4558--4562},
  doi       = {10.21437/Interspeech.2024-2033},
  issn      = {2958-1796},
}

@inproceedings{BERTScore,
  author       = {Tianyi Zhang and
                  Varsha Kishore and
                  Felix Wu and
                  Kilian Q. Weinberger and
                  Yoav Artzi},
  title        = {BERTScore: Evaluating Text Generation with {BERT}},
  booktitle    = {8th International Conference on Learning Representations, {ICLR},
                  Addis Ababa, Ethiopia, April 26-30},
  year         = {2020},
  url          = {https://openreview.net/forum?id=SkeHuCVFDr},
  bibsource    = {dblp computer science bibliography, https://dblp.org}
}

\end{document}